%% file: main.tex
\documentclass[conference]{IEEEtran}
\IEEEoverridecommandlockouts
\usepackage{cite}
\usepackage[font=small,labelfont=bf]{caption}
\usepackage{pgfplots}
\pgfplotsset{compat=1.17} %
\usepackage{amsmath,amssymb,amsfonts}
\usepackage{algorithmic}
\usepackage{graphicx}
\usepackage{textcomp}
\usepackage{booktabs}  
\usepackage{multirow}  
\usepackage{amsmath}   
\usepackage{xcolor}
\def\BibTeX{{\rm B\kern-.05em{\sc i\kern-.025em b}\kern-.08em
    T\kern-.1667em\lower.7ex\hbox{E}\kern-.125emX}}
\usepackage{graphicx}   
\usepackage{subcaption} 
\usepackage{float}      
    
\begin{document}

\title{Domain Generalization in Autonomous Driving: Evaluating YOLOv8s, RT-DETR, and YOLO-NAS with the ROAD-Almaty Dataset\\
}


\author{
     Madiyar Alimov\IEEEauthorrefmark{1}, Temirlan Meiramkhanov\IEEEauthorrefmark{2} \\

    \IEEEauthorblockA{\IEEEauthorrefmark{2}Department of Computer Technologies and Cybersecurity, International IT University, Almaty, Kazakhstan \\
    Email: 29804@iitu.edu.kz}
    \IEEEauthorblockA{\IEEEauthorrefmark{1}Department of Computational and Data Sciences, Astana IT University, Astana, Kazakhstan \\
    Email: 242813@astanait.edu.kz}

}

\maketitle

\begin{abstract}

This study investigates the domain generalization capabilities of three state-of-the-art object detection models—YOLOv8s, RT-DETR, and YOLO-NAS—within the unique driving environment of Kazakhstan. Utilizing the newly constructed ROAD-Almaty dataset, which encompasses diverse weather, lighting, and traffic conditions, we evaluated the models’ performance without any retraining. Quantitative analysis revealed that RT-DETR achieved an average F1-score of 0.672 at IoU=0.5, outperforming YOLOv8s (0.458) and YOLO-NAS (0.526) by approximately 46\% and 27\%, respectively. Additionally, all models exhibited significant performance declines at higher IoU thresholds (e.g., a drop of approximately 20\% when increasing IoU from 0.5 to 0.75) and under challenging environmental conditions, such as heavy snowfall and low-light scenarios. These findings underscore the necessity for geographically diverse training datasets and the implementation of specialized domain adaptation techniques to enhance the reliability of autonomous vehicle detection systems globally. This research contributes to the understanding of domain generalization challenges in autonomous driving, particularly in underrepresented regions.

\end{abstract}


\input{sections/1_intor}
\input{sections/2_rel_work}

\input{sections/3_dataset}

\input{sections/4_results_and_discussion}
\input{sections/5_conclusion}

\bibliographystyle{ieeetr}
\bibliography{references}

\end{document}

%% file: sections/1_intor.tex
\section{INTRODUCTION}
The increasing complexity of urban transportation systems and growing safety concerns have driven the rapid development of advanced vehicle detection and tracking technologies. Within Intelligent Transportation Systems (ITS), accurate and efficient object detection is essential for enhancing road safety, reducing congestion, and improving traffic management. In recent years, deep learning-based methods, particularly Convolutional Neural Networks (CNNs), have dramatically improved object detection accuracy and speed \cite{simonyan2014very}. Landmark approaches such as Fast R-CNN \cite{girshick2015fast} established a solid foundation, while the YOLO (You Only Look Once) family introduced real-time detection with competitive accuracy
\cite{redmon2016you} \cite{bochkovskiy2020yolov4} \cite{redmon2018yolov3}

Despite these technological gains, most object detection models are evaluated on datasets with relatively homogenous conditions, limiting our understanding of how they perform under more challenging, diverse scenarios. In practice, environmental factors—ranging from extreme weather and poor lighting to varied roadway infrastructures—can significantly degrade model performance when applied to new, underrepresented domains \cite{torralba2011unbiased}. This issue is particularly pronounced in regions like Kazakhstan, where unique weather patterns, lighting conditions, and traffic behaviors differ markedly from commonly studied locales \cite{beery2018recognition}. The resulting domain shift raises critical questions about the robustness and adaptability of state-of-the-art models that have not been tested extensively in such settings.

To address this gap, our research examines how three leading pre-trained object detection models—YOLOv8s [8], RT-DETR [9], and YOLO-NAS [10]—generalize to vehicle detection tasks in Kazakhstan’s distinct driving environment without additional retraining. Specifically, we ask:

\begin{enumerate}
    \item How well do these models perform under the diverse weather, lighting, and traffic conditions encountered in Kazakhstan?
    \item Which model demonstrates the strongest capacity for domain generalization across previously unseen scenarios?
\end{enumerate}

We hypothesize that significant performance variability will emerge due to domain shift. Understanding and mitigating these shifts is not only a theoretical exercise but also a critical step towards ensuring that autonomous vehicles can be safely and efficiently deployed in diverse global markets, thereby expanding their practical impact. By providing empirical insights into model robustness in this less-explored environment, our study aims to advance the field toward more universally reliable autonomous vehicle detection systems, ultimately supporting safer and more efficient transportation worldwide.

%% file: sections/2_rel_work.tex
\section{RELATED WORK}
\subsection{Advances in Object Detection for Autonomous Driving}
Deep learning-based methods have substantially improved the accuracy and efficiency of object detection in autonomous driving, enabling vehicles to better understand their surroundings \cite{simonyan2014very}. Early CNN-based detectors, such as Fast R-CNN \cite{girshick2015fast} and Faster R-CNN  \cite{ren2016faster}, offered accuracy gains but often struggled with real-time performance constraints. The YOLO series \cite{redmon2016you}, \cite{bochkovskiy2020yolov4}, \cite{redmon2018yolov3} addressed these limitations by reframing detection as a single-step regression problem, achieving a balance between speed and accuracy that suits deployment in Intelligent Transportation Systems.

\subsection{The Challenge of Domain Shift}
While these advances are noteworthy, they often assume that the model’s training and testing environments share similar characteristics. In practice, autonomous vehicles operate under diverse conditions, and even state-of-the-art models can falter when exposed to new domains with different weather patterns, lighting, road designs, and traffic behaviors \cite{torralba2011unbiased}. This domain shift problem is particularly acute when deploying models in regions like Kazakhstan, where environmental conditions can deviate markedly from the scenarios captured in widely used datasets \cite{beery2018recognition}.

\subsection{Existing Datasets and Their Geographic Limitations}

Numerous datasets have been instrumental in advancing object detection for autonomous driving, yet most focus on a narrow range of geographic and environmental conditions. For example, KITTI \cite{Geiger2013IJRR} and KITTI-360 \cite{Liao2022PAMI} facilitated progress in 3D detection and scene understanding, while ApolloScape \cite{10.1609/aaai.v33i01.33016120}, PIE \cite{9008118}, nuScenes \cite{nuscenes2019}, and WAYMO \cite{9709630} introduced richer data modalities and more varied urban settings—primarily in North America, Europe, or East Asia. Likewise, Argoverse \cite{Argoverse} and Lyft Level 5 \cite{Houston2020OneTA} supported complex trajectory prediction tasks but still reflected relatively similar road infrastructures and cultural driving patterns.

Although more recent datasets have attempted to broaden environmental coverage—such as BDD100K \cite{yu2020bdd100k}, which includes a range of weather and lighting conditions, or synthetic datasets like SHIFT \cite{9880320} and Virtual KITTI 2 \cite{cabon2020virtualkitti2}, which allow controlled parameter variations—these resources nonetheless remain limited in geographic scope. The unique climates, infrastructures, and traffic behaviors characteristic of regions like Central Asia have not been adequately captured. This gap hinders the development of universally robust models and underscores the need for datasets that truly represent the global diversity of driving environments, including locales like Kazakhstan.

\subsection{Approaches to Domain Adaptation and Generalization}

Researchers have explored various strategies to mitigate domain shift. Unsupervised Domain Adaptation (UDA) techniques \cite{long2016unsupervised} leverage unlabeled target domain data to refine feature extraction, while adversarial training \cite{tzeng2017adversarial} encourages models to learn domain-invariant representations. Domain-Adversarial Neural Networks (DANN) \cite{ghifary2015domain} promote features that are indistinguishable between source and target domains, enhancing cross-domain generalization. Despite these methodological advances, most studies focus on adapting between similar urban environments or weather conditions, rather than addressing the unique challenges posed by less-explored geographical regions.

\subsection{State-of-the-Art Models and Unanswered Questions}

Recent architectures like YOLOv8 \cite{ultralytics2023yolov8}, RT-DETR \cite{ultralytics2023rtdetr}, and YOLO-NAS \cite{ultralytics2023yolon} have set new performance benchmarks. While these references currently cite official documentation rather than peer-reviewed papers, they are recognized and widely used within the computer vision community. Future peer-reviewed publications on these models would further strengthen their scholarly basis.

YOLOv8 refines the YOLO paradigm with architectural tweaks for speed and accuracy; RT-DETR leverages transformer-based attention mechanisms; and YOLO-NAS uses Neural Architecture Search to optimize configurations for diverse scenarios. However, their performance under true out-of-distribution conditions—such as those found in Kazakhstan—remains unclear.

\subsection{Identifying the Research Gap}

In sum, the literature demonstrates impressive gains in object detection within known domains and controlled variations, as well as promising domain adaptation techniques aimed at bridging differences between datasets. Still, an essential gap persists: we lack empirical evidence on how top-performing, pre-trained models fare when confronted with markedly different geographical and environmental settings. The absence of studies evaluating these models in Central Asia underscores the need for targeted exploration.

This study addresses that gap by evaluating YOLOv8s, RT-DETR, and YOLO-NAS in the Kazakhstani context without retraining. By doing so, we provide insight into their innate generalization capabilities and the extent to which domain shift affects their performance. Understanding these limitations is a crucial step toward developing more globally reliable autonomous driving systems.

%% file: sections/3_dataset.tex
\section{METHODOLOGY}
\subsection{Data Collection and Preparation}
To evaluate the generalization performance of our chosen object detection models (YOLOv8s, RT-DETR, YOLO-NAS) under diverse environmental conditions in Kazakhstan, we constructed the ROAD-Almaty dataset. The data were collected using a single instrumented vehicle equipped with a HYBRID-UNO-SPORT-WiFi dashcam. Recordings took place over multiple sessions in Almaty, Kazakhstan, encompassing variations in weather (clear, rainy, foggy, and cloudy) and time-of-day (daytime and nighttime) to reflect the complex driving scenarios prevalent in the region.

Each recording lasted one minute, captured at 30 frames per second with a 1920×1080 resolution. To balance comprehensive coverage with manageable annotation effort, we sampled frames at a rate of 10 fps, resulting in a total of 1,844 annotated images. Among these annotated instances, we identified  9,286 cars, 320 buses, 103 trucks, 34 motorcycles, and 455 pedestrians. This distribution reflects the variety of road users encountered in Almaty’s traffic environment. These images were also selected to represent a spectrum of traffic conditions, road types (highways, narrow streets), and visibility challenges (glare, low-light, and partial occlusions).

\subsection{Annotation Protocol and Ethical Considerations}

Annotations were performed using Roboflow, focusing on objects critical for autonomous driving: cars, buses, trucks, motorcycles, and pedestrians. A team of trained annotators applied bounding boxes and class labels to each frame. To ensure annotation quality, a two-stage review was implemented: initial annotations were checked and refined by experienced annotators. This iterative process aimed at minimizing mislabeling, especially in low-visibility frames.

Throughout data collection and annotation, we followed standard ethical guidelines by ensuring that no personally identifying features (e.g., license plates, facial details of pedestrians) were retained in a manner that violates privacy. Where possible, identifiable elements were obscured or avoided in final datasets.

\subsubsection{Data Splits and Experimental Setup}
To rigorously assess model performance, we divided the ROAD-Almaty dataset into training, validation, and testing sets. Although models under evaluation were pre-trained on diverse source domains (e.g., COCO, YOLO’s proprietary sets), they received no retraining on our data. Instead, our methodology involves evaluating these off-the-shelf models directly on the test portion of ROAD-Almaty to measure true out-of-distribution generalization capability.

In particular, we reserved approximately 20\% of frames for testing. These test frames were selected to maximize environmental variability, ensuring that the testing phase included challenging weather (e.g., heavy rain, low visibility fog) and lighting conditions (twilight or nighttime). The remaining portion of the dataset was retained as a reference or future adaptation resource and was not used to train or fine-tune the models for this study.

\subsubsection{Evaluation Metrics and Tools}

To quantify performance, we employed standard object detection metrics, including Intersection over Union (IoU) at thresholds of 0.5 and 0.75 and F1-scores derived from precision and recall. These metrics were computed using standard evaluation scripts aligned with the COCO benchmarking format. By maintaining consistent evaluation parameters, our methodology allows a fair comparison between YOLOv8s, RT-DETR, and YOLO-NAS under identical conditions.

This methodological structure—encompassing diverse data collection, rigorous annotation, careful dataset splitting, and standardized evaluation metrics—enables us to systematically investigate which pre-trained model best generalizes to the unique driving environment in Kazakhstan, thus directly addressing the research questions posed in this study.

\subsection{Sample Images from ROAD-Almaty Dataset}

\begin{figure*}[t] 
    \centering
    \setlength{\tabcolsep}{2pt} 
    \renewcommand{\arraystretch}{1} 

    \begin{tabular}{ccc} 
        \includegraphics[width=0.3\textwidth]{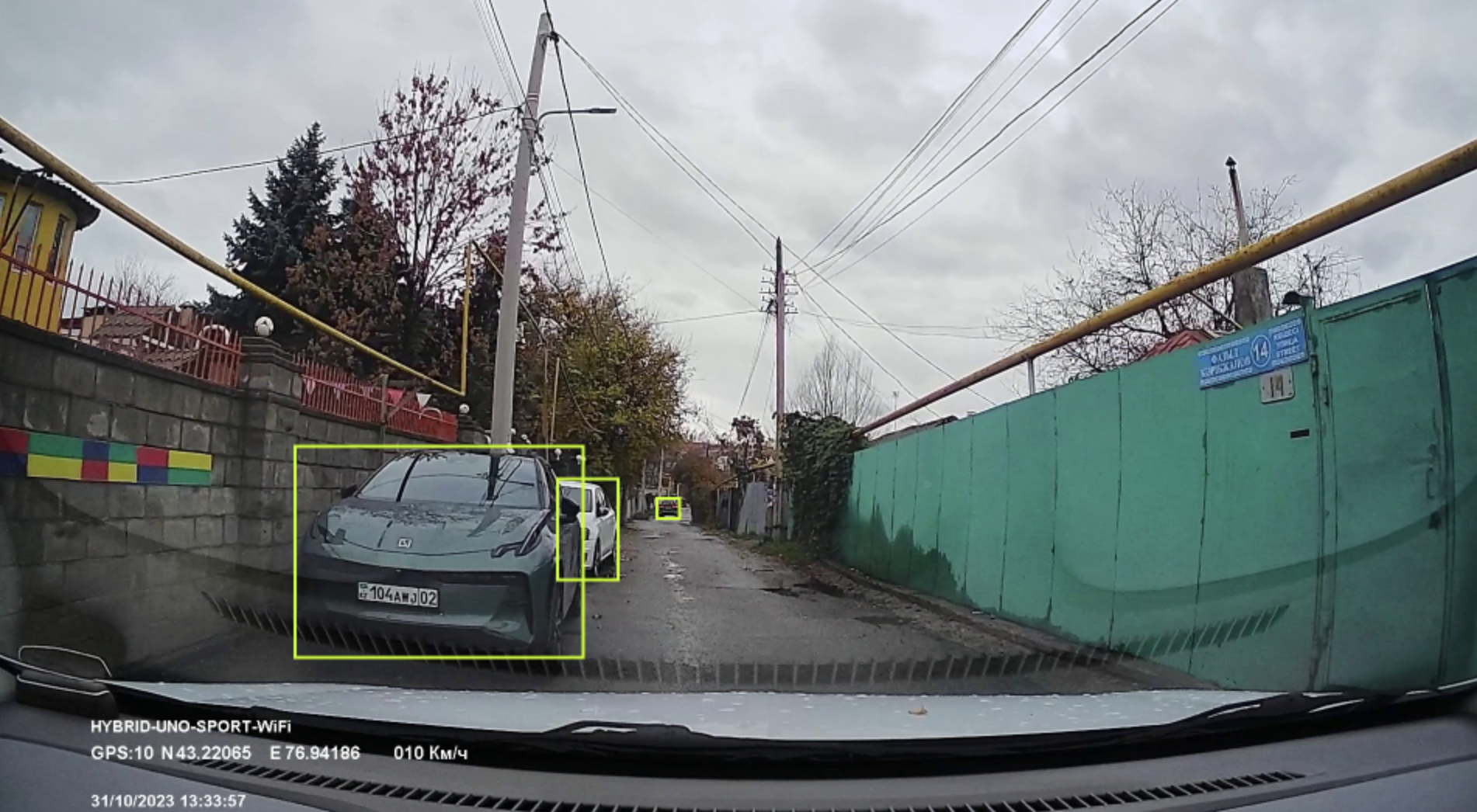} &
        \includegraphics[width=0.3\textwidth]{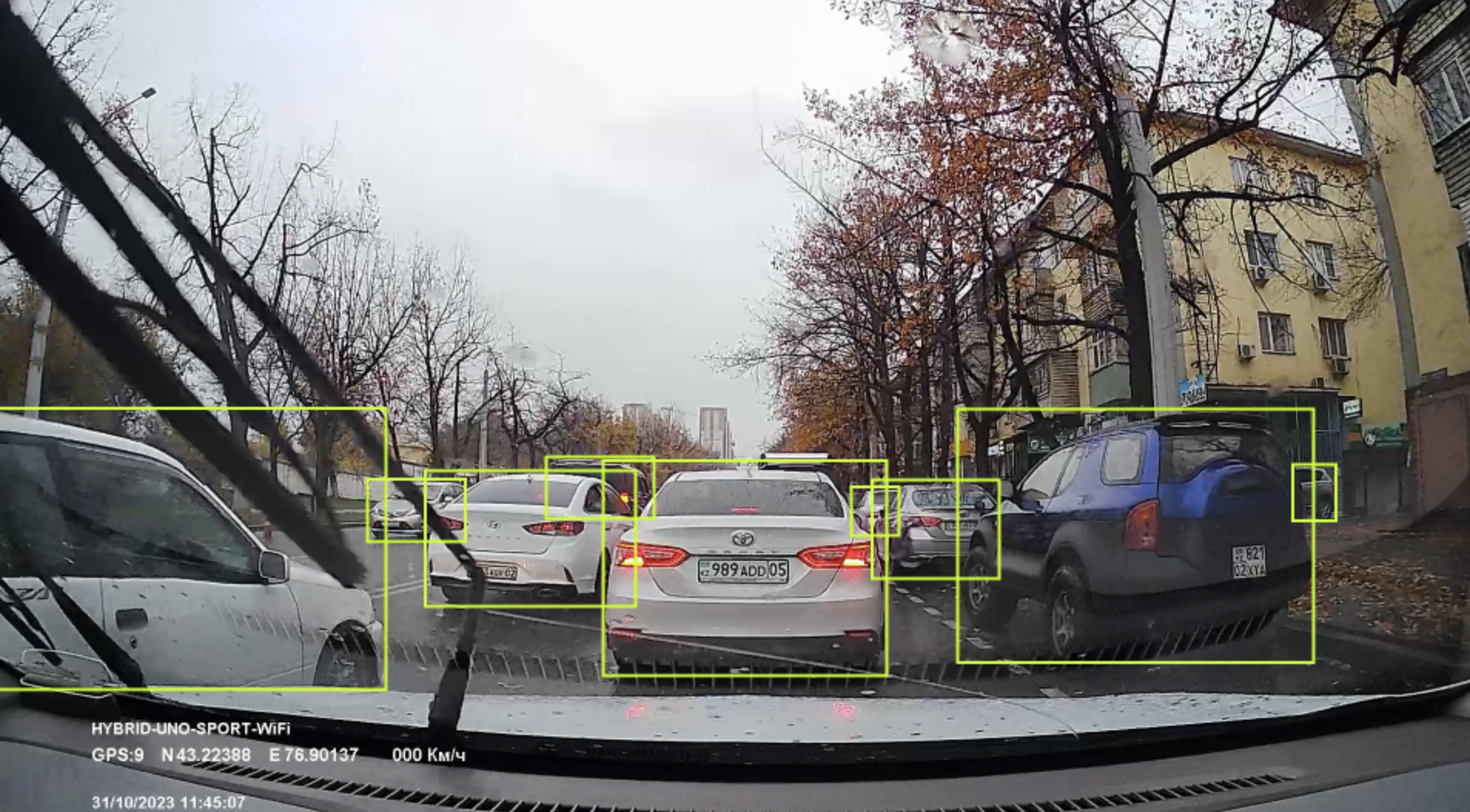} &
        \includegraphics[width=0.3\textwidth]{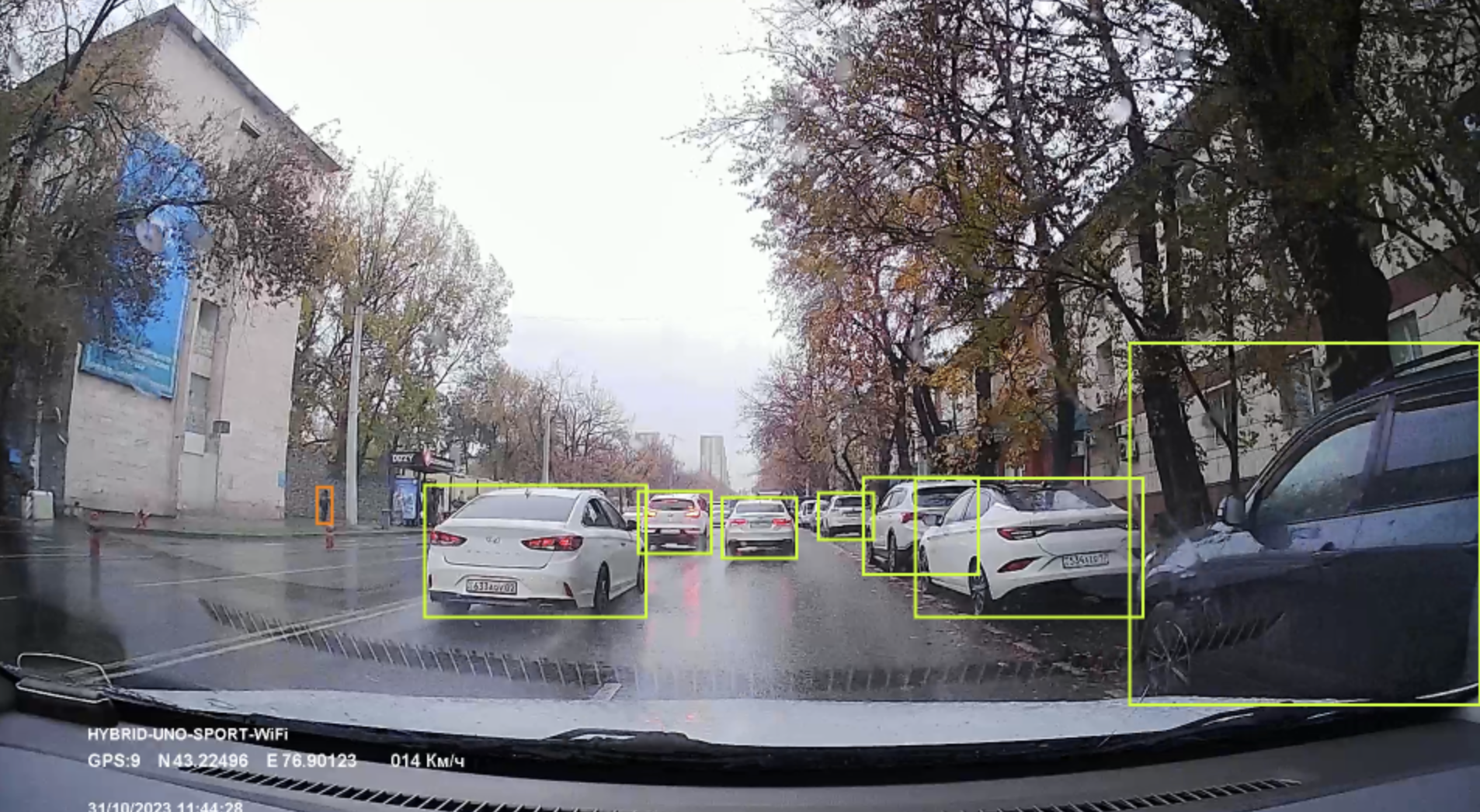} \\
        \includegraphics[width=0.3\textwidth]{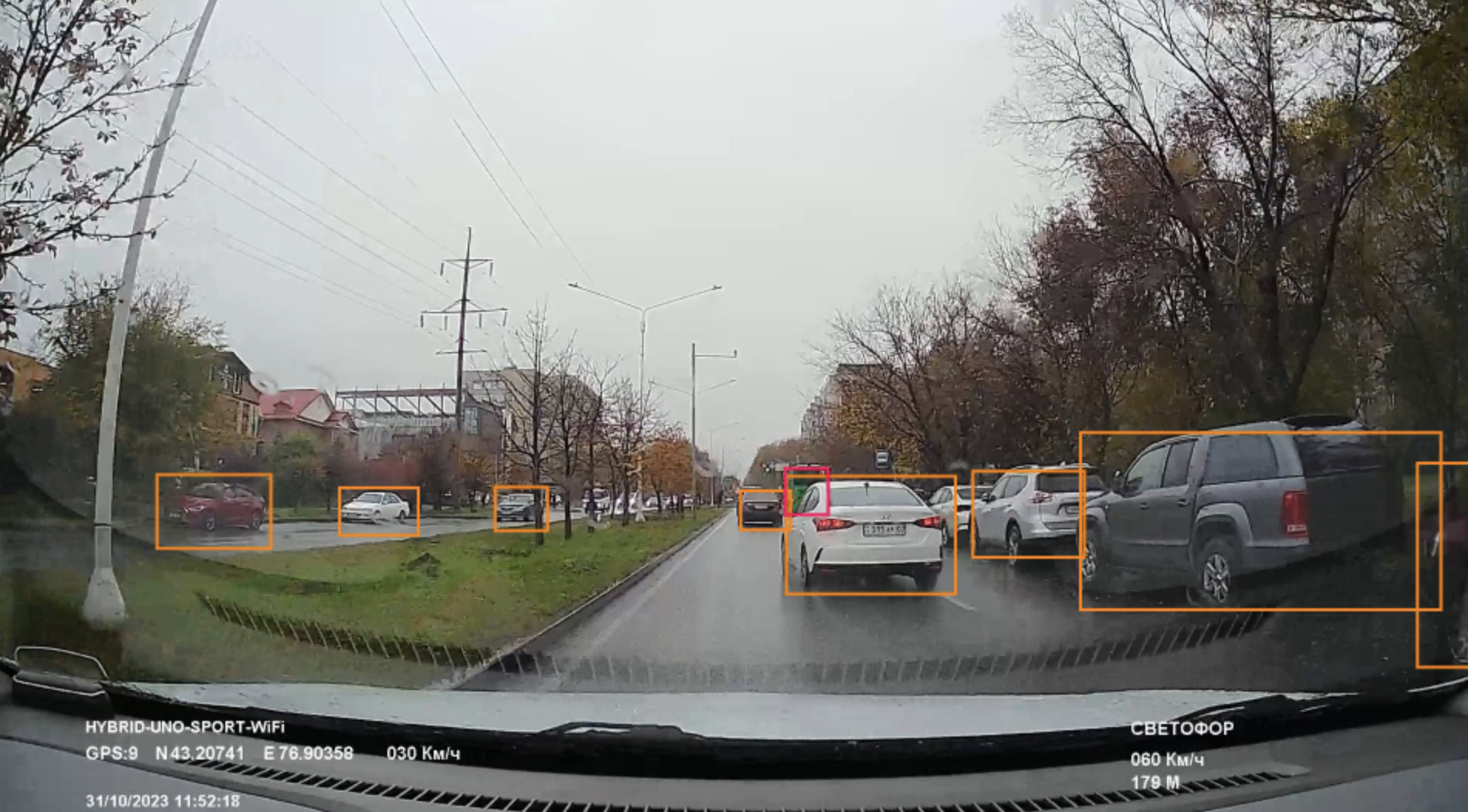} &
        \includegraphics[width=0.3\textwidth]{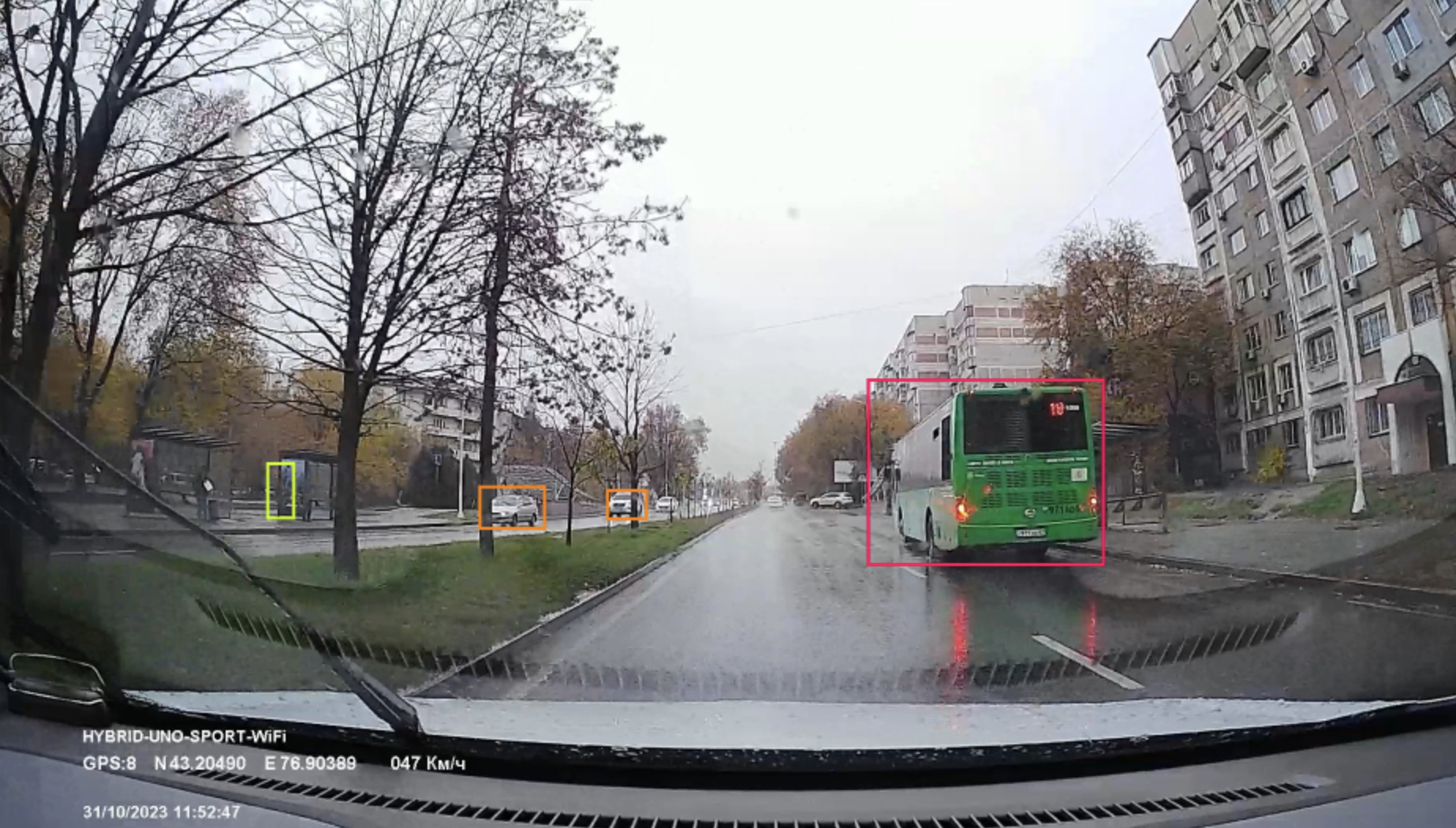} &
        \includegraphics[width=0.3\textwidth]{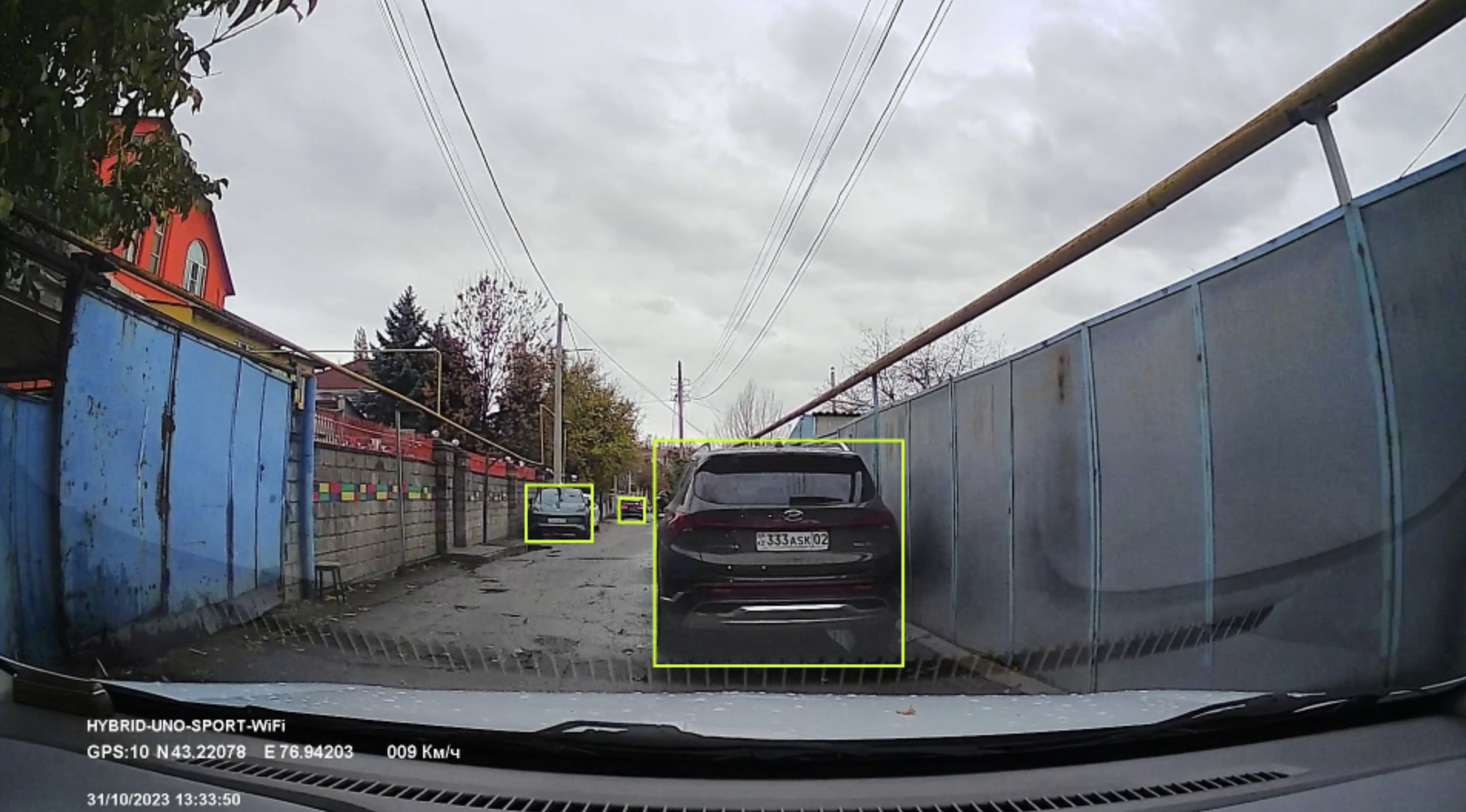} \\
        \includegraphics[width=0.3\textwidth]{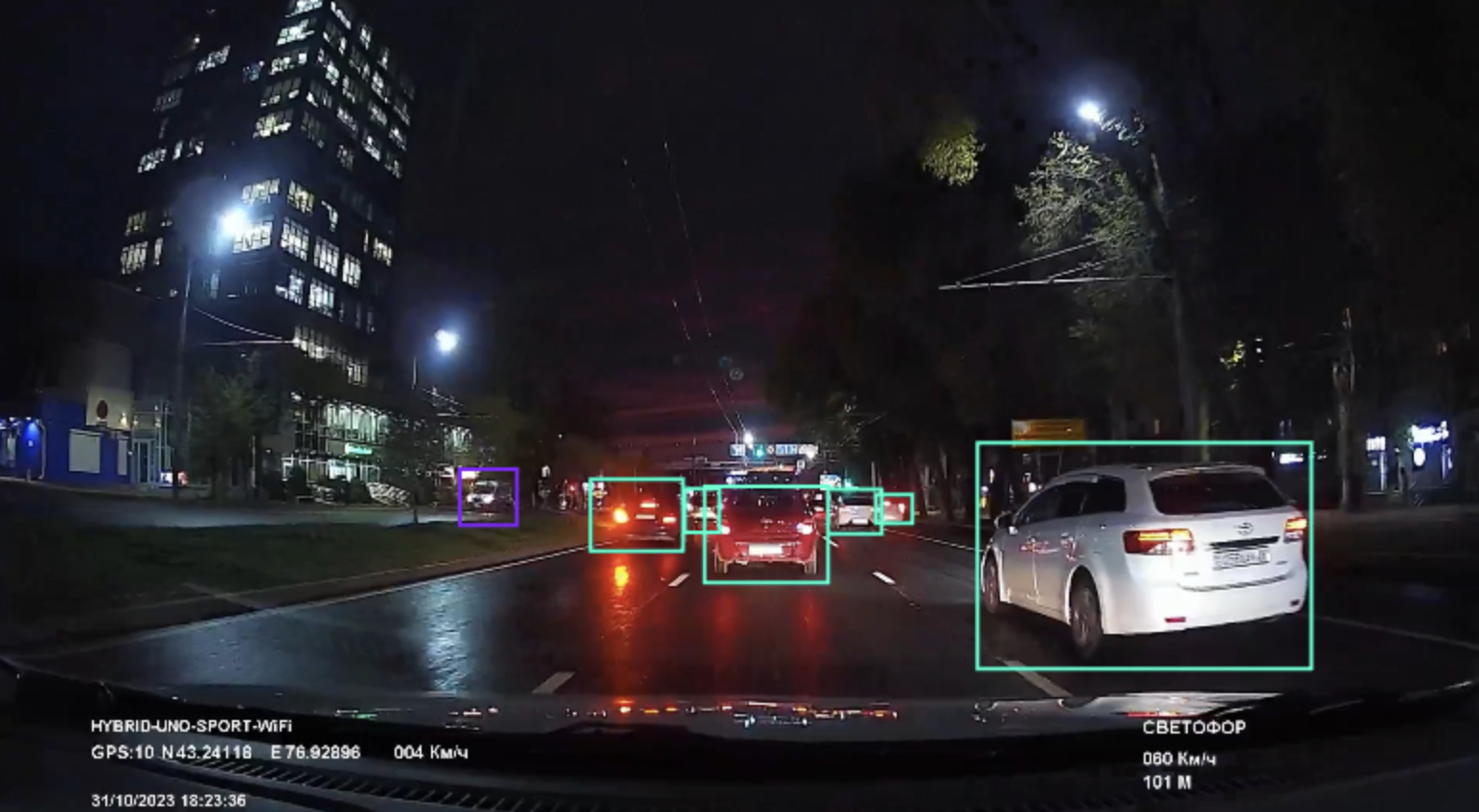} &
        \includegraphics[width=0.3\textwidth]{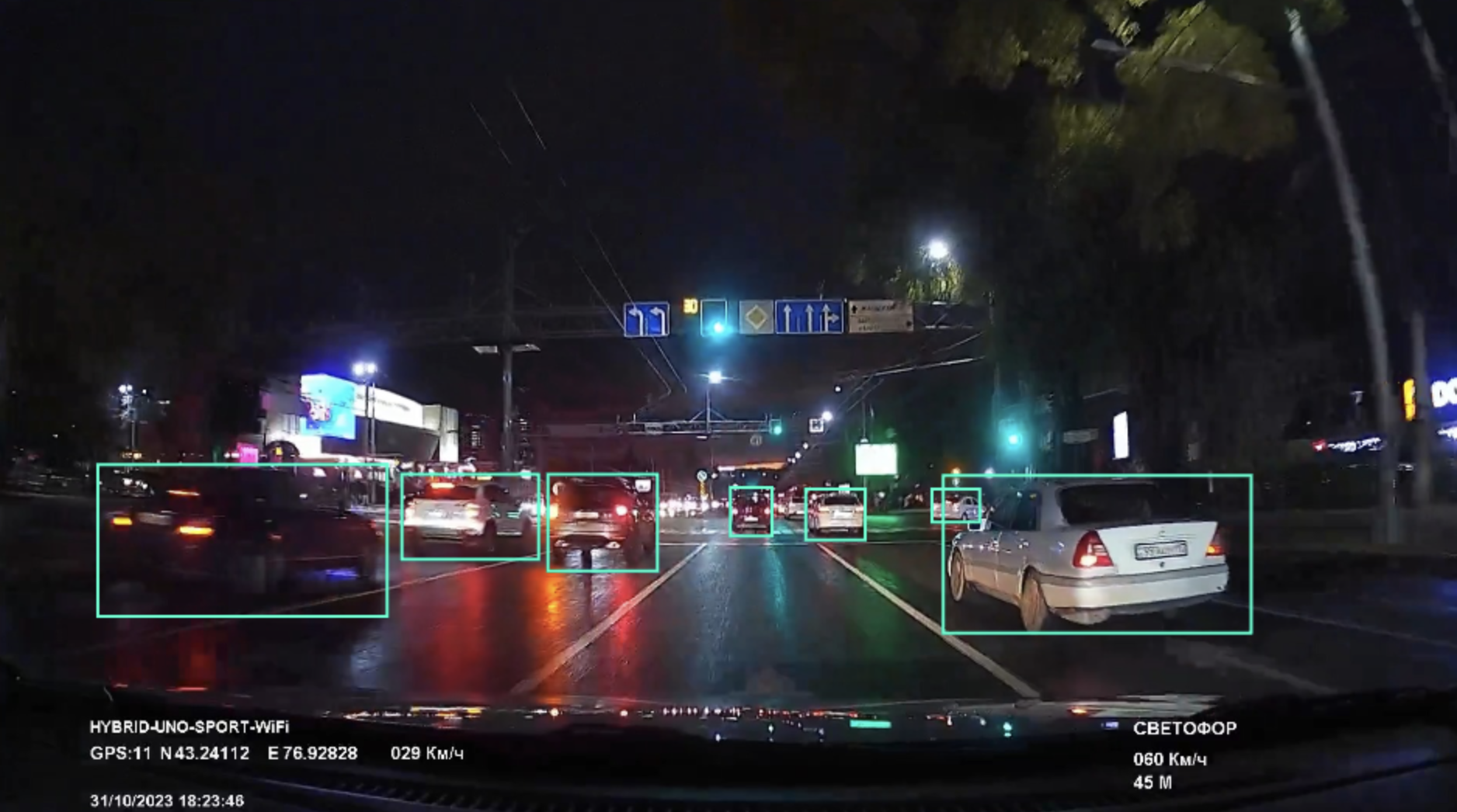} &
        \includegraphics[width=0.3\textwidth]{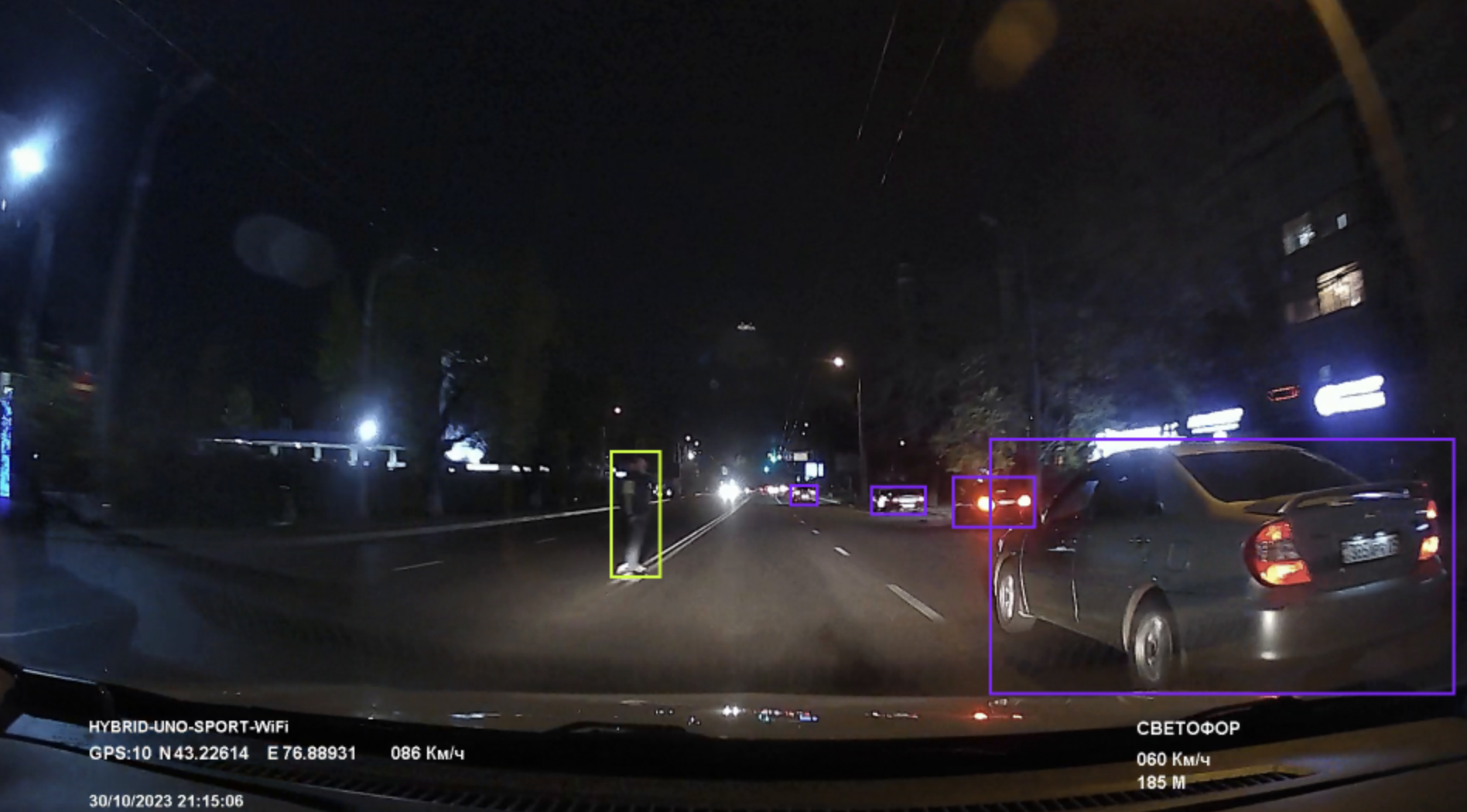} \\
    \end{tabular}

    \caption{Collage of sample frames from the ROAD-Almaty dataset under varying weather and lighting conditions.}
    \label{fig:road_almaty_collage}
\end{figure*}

To illustrate the diversity of weather and lighting conditions in our dataset, Figure~\ref{fig:road_almaty_collage} presents a grid of representative frames.

%% file: sections/4_results_and_discussion.tex
\section{RESULTS}
We evaluated the generalization capabilities of three state-of-the-art object detection models—YOLOv8s, RT-DETR, and YOLO-NAS—using the ROAD-Almaty dataset collected in Kazakhstan. The evaluation was conducted without retraining or fine-tuning, thereby allowing a direct assessment of how well these models, initially trained on widely used datasets, handle out-of-distribution conditions.
\subsection{Quantitative Performance Analysis}
Table I presents the detailed IoU and F1-scores for each sequence. To further illustrate these differences, Figure 2 shows a grouped bar chart of the F1-scores at IoU=0.5 across the five test sequences. This visualization clearly indicates that RT-DETR consistently achieves higher F1-scores than both YOLOv8s and YOLO-NAS, especially in Seq1 and Seq4, while sequences with more challenging conditions (e.g., Seq3) see all models performing poorly, albeit RT-DETR remains relatively superior.

\begin{table*}[t!]
\centering
\caption{F1 and IoU Scores Across Sequences for Different Models (Average Values)}
\begin{tabular}{llcccccc}
\toprule
Model      & IoU Thresh & Seq1 (IoU/F1) & Seq2 (IoU/F1) & Seq3 (IoU/F1) & Seq4 (IoU/F1) & Seq5 (IoU/F1) & Average (IoU/F1) \\ \midrule
NAS        & 0.5        & 0.878/0.832   & 0.584/0.324   & 0.366/0.315   & 0.878/0.832   & 0.584/0.324   & 0.658/0.526      \\
NAS        & 0.75       & 0.895/0.689   & 0.454/0.225   & 0.226/0.174   & 0.895/0.689   & 0.454/0.225   & 0.585/0.400      \\
RT-DETR    & 0.5        & 0.890/0.900   & 0.692/0.517   & 0.476/0.527   & 0.890/0.900   & 0.692/0.517   & 0.728/0.672      \\
RT-DETR    & 0.75       & 0.910/0.820   & 0.577/0.328   & 0.368/0.327   & 0.910/0.820   & 0.577/0.328   & 0.668/0.525      \\
YOLOv8s    & 0.5        & 0.876/0.766   & 0.517/0.255   & 0.307/0.247   & 0.876/0.766   & 0.517/0.255   & 0.619/0.458      \\
YOLOv8s    & 0.75       & 0.895/0.587   & 0.378/0.159   & 0.197/0.133   & 0.895/0.587   & 0.378/0.159   & 0.549/0.325      \\ \bottomrule
\end{tabular}
\end{table*}

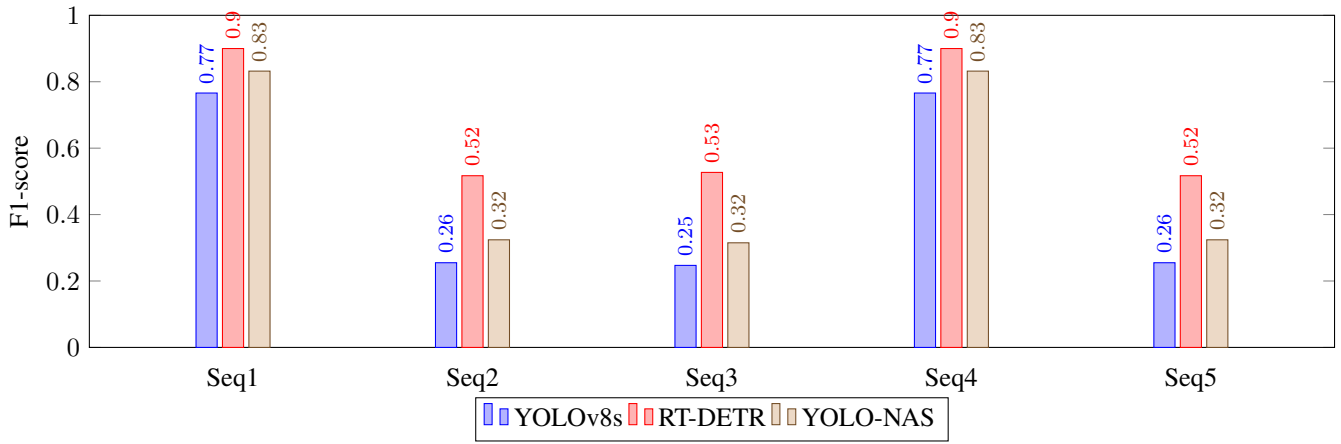
\begin{figure*}[t]
    \centering
    \begin{tikzpicture}
        \begin{axis}[
            ybar,
            bar width=8pt,
            width=\textwidth,
            height=6cm,
            enlarge x limits=0.15,
            legend style={at={(0.5,-0.15)}, anchor=north, legend columns=-1},
            symbolic x coords={Seq1, Seq2, Seq3, Seq4, Seq5},
            xtick=data,
            ymin=0, ymax=1,
            ylabel={F1-score},
            xlabel={Test Sequences},
            nodes near coords,
            every node near coord/.append style={font=\footnotesize, rotate=90, anchor=west},
            xtick style={draw=none},
            ]
            \addplot coordinates {(Seq1,0.766) (Seq2,0.255) (Seq3,0.247) (Seq4,0.766) (Seq5,0.255)};
            \addplot coordinates {(Seq1,0.900) (Seq2,0.517) (Seq3,0.527) (Seq4,0.900) (Seq5,0.517)};
            \addplot coordinates {(Seq1,0.832) (Seq2,0.324) (Seq3,0.315) (Seq4,0.832) (Seq5,0.324)};
            \legend{YOLOv8s, RT-DETR, YOLO-NAS}
        \end{axis}
    \end{tikzpicture}
    \caption{Comparative F1-scores at IoU=0.5 across five test sequences (Seq1–Seq5) for YOLOv8s, RT-DETR, and YOLO-NAS. Each group of three bars represents a single sequence, illustrating how performance varies with changing environmental conditions and how RT-DETR consistently outperforms the YOLO-based models.}
    \label{fig:model_performance}
\end{figure*}

\subsection{Model Comparisons}

\subsubsection{RT-DETR Outperforms YOLO-Based Models} 
Across both IoU thresholds, RT-DETR consistently demonstrates superior average IoU and F1-scores. At a 0.5 IoU threshold, RT-DETR achieves an average F1-score of approximately 0.672, outperforming YOLOv8s (0.458) and YOLO-NAS (0.526). This suggests that the transformer-based approach in RT-DETR may offer more robust feature extraction and localization when confronted with unfamiliar conditions.

\subsubsection{Performance Drop at Higher IoU Thresholds}

As expected, increasing the IoU threshold from 0.5 to 0.75 reduces detection performance across all models. YOLO-NAS and YOLOv8s, which rely on CNN-based structures and known architectural heuristics, exhibit more pronounced drops in both IoU and F1. RT-DETR, while also affected, maintains relatively better performance. This indicates that stricter localization criteria amplify the domain shift challenges, particularly for models originally tuned to more conventional conditions.

\subsubsection{Variability Across Sequences}

There is noticeable variability in model performance across different sequences. Some sequences, likely captured under challenging lighting or weather conditions, result in substantial performance degradation. This variability underscores the complexity of Kazakhstani road scenarios and highlights the necessity of evaluating models in such underrepresented domains. Models that generalize well across varied sequences are more promising candidates for widespread deployment.

\subsection{Implications for Domain Generalization}

These findings confirm our hypothesis that domain shift negatively impacts detection accuracy. While RT-DETR appears more resilient, it too experiences performance declines relative to results reported in more familiar datasets. The results suggest that even top-performing models have significant room for improvement when dealing with geographically distinct and environmentally complex regions. This underscores the need for specialized domain adaptation techniques and more diverse training data to achieve consistent performance in real-world conditions.

\section{Discussion}
The results presented in Section IV highlight the complexity of deploying pre-trained object detection models in geographically and environmentally distinct regions. While RT-DETR demonstrated the highest overall performance, particularly at an IoU threshold of 0.5, even this model’s accuracy declined substantially under more stringent localization criteria (IoU = 0.75) and challenging environmental conditions. YOLOv8s and YOLO-NAS, despite their strong performance in more conventional datasets, struggled to maintain consistency when confronted with Kazakhstani driving scenarios.

\subsection{Comparison with Previous Studies}
Our findings align with prior research indicating that models trained on canonical datasets can experience significant performance degradation when tested in out-of-distribution conditions \cite{torralba2011unbiased}, \cite{beery2018recognition}. Similar to studies highlighting the importance of environmental diversity (e.g., BDD100K \cite{yu2020bdd100k}) and synthetic variations (e.g., SHIFT \cite{9880320}), our work reaffirms the need for broader geographic and climatic coverage in training data. However, unlike these earlier studies, our evaluation explicitly targets an underrepresented region—Kazakhstan—thereby extending the understanding of domain shift to a locale not extensively documented in existing literature.

Furthermore, while methods like UDA \cite{long2016unsupervised}, adversarial training \cite{tzeng2017adversarial}, and DANN \cite{ghifary2015domain} have shown promise in narrowing the domain gap, most prior work has focused on either weather or time-of-day variations within well-studied urban environments. Our results suggest that these techniques may require adaptation or further refinement to handle more drastic domain shifts, such as those introduced by distinct road infrastructures, traffic patterns, and cultural driving behaviors found in Central Asia.

\subsection{Limitations and Future Directions}
A key limitation of our study is that we did not apply any domain adaptation techniques to improve model performance. While this allowed for a pure assessment of the models’ innate generalization capabilities, future work could explore integrating UDA or DANN-based approaches tailored to the distinct conditions in Kazakhstan. Additionally, our dataset, although diverse within the region, is smaller than large-scale benchmarks like the Waymo Open Dataset \cite{9709630}. Expanding the dataset’s temporal and spatial coverage, as well as incorporating multiple cities and additional environmental factors, would create a richer testbed for understanding domain shifts. Exploring sensor fusion or additional modalities (e.g., LiDAR, radar) may further enhance performance in challenging scenarios.

\subsection{Practical Implications}
From an application standpoint, the results underscore the importance of geographic diversity in training and evaluating autonomous vehicle models. Before deploying AVs in new regions, it may be necessary for manufacturers and policymakers to incorporate local data collection or apply advanced adaptation strategies. By doing so, they can ensure that AV systems maintain reliable performance and safety standards, even under conditions markedly different from those in their original training environments.

\subsection{Contribution to the Field}
Our study contributes to the emerging literature on domain generalization by offering empirical evidence of how state-of-the-art object detection models perform in a less-explored domain. This research highlights that even top-performing models may falter without proper domain adaptation. By identifying these limitations, we pave the way for the development of more robust, global-scale solutions that can confidently operate in a wider array of environmental and geographic contexts.

%% file: sections/5_conclusion.tex
\section{Conclusion}

In this study, we evaluated the performance of three pre-trained object detection models—YOLOv8s, RT-DETR, and YOLO-NAS—on the ROAD-Almaty dataset, representing the unique driving conditions of Kazakhstan. Our findings demonstrate that even state-of-the-art models, initially trained on widely used datasets, can experience significant performance degradation when applied to geographically distinct and environmentally challenging scenarios. RT-DETR exhibited comparatively better resilience, yet it, too, suffered declines under stringent localization criteria and adverse conditions.

These results underscore the pressing need for more geographically diverse training datasets and the integration of specialized domain adaptation techniques. By empirically revealing the limitations of leading models in previously underrepresented contexts, this work contributes to a growing understanding of domain generalization challenges in autonomous driving. Ultimately, addressing these challenges—through strategies such as targeted domain adaptation frameworks, synthetic data augmentation, and sensor fusion—will enable safer, more reliable global deployments of autonomous vehicles, better reflecting the complex realities of driving conditions worldwide.

%% file: main.bbl
\begin{thebibliography}{10}

\bibitem{simonyan2014very}
K.~Simonyan and A.~Zisserman, ``Very deep convolutional networks for large-scale image recognition,'' {\em arXiv preprint arXiv:1409.1556}, 2014.

\bibitem{girshick2015fast}
R.~Girshick, ``Fast r-cnn,'' in {\em Proceedings of the IEEE International Conference on Computer Vision}, pp.~1440--1448, 2015.

\bibitem{redmon2016you}
J.~Redmon, S.~Divvala, R.~Girshick, and A.~Farhadi, ``You only look once: Unified, real-time object detection,'' in {\em Proceedings of the IEEE Conference on Computer Vision and Pattern Recognition}, pp.~779--788, 2016.

\bibitem{bochkovskiy2020yolov4}
A.~Bochkovskiy, C.-Y. Wang, and H.-Y.~M. Liao, ``Yolov4: Optimal speed and accuracy of object detection,'' {\em arXiv preprint arXiv:2004.10934}, 2020.

\bibitem{redmon2018yolov3}
J.~Redmon, ``Yolov3: An incremental improvement,'' {\em arXiv preprint arXiv:1804.02767}, 2018.

\bibitem{torralba2011unbiased}
A.~Torralba and A.~A. Efros, ``Unbiased look at dataset bias,'' in {\em Proceedings of the IEEE Conference on Computer Vision and Pattern Recognition (CVPR)}, pp.~1521--1528, June 2011.

\bibitem{beery2018recognition}
S.~Beery, G.~Van~Horn, and P.~Perona, ``Recognition in terra incognita,'' in {\em Proceedings of the European Conference on Computer Vision (ECCV)}, pp.~456--473, 2018.

\bibitem{ren2016faster}
S.~Ren, K.~He, R.~Girshick, and J.~Sun, ``Faster r-cnn: Towards real-time object detection with region proposal networks,'' {\em IEEE transactions on pattern analysis and machine intelligence}, vol.~39, no.~6, pp.~1137--1149, 2016.

\bibitem{Geiger2013IJRR}
A.~Geiger, P.~Lenz, C.~Stiller, and R.~Urtasun, ``Vision meets robotics: The kitti dataset,'' {\em International Journal of Robotics Research (IJRR)}, 2013.

\bibitem{Liao2022PAMI}
Y.~Liao, J.~Xie, and A.~Geiger, ``{KITTI}-360: A novel dataset and benchmarks for urban scene understanding in 2d and 3d,'' {\em Pattern Analysis and Machine Intelligence (PAMI)}, 2022.

\bibitem{10.1609/aaai.v33i01.33016120}
Y.~Ma, X.~Zhu, S.~Zhang, R.~Yang, W.~Wang, and D.~Manocha, ``Trafficpredict: Trajectory prediction for heterogeneous traffic-agents,'' in {\em Proceedings of the AAAI Conference on Artificial Intelligence}, 2019.

\bibitem{9008118}
A.~Rasouli, I.~Kotseruba, T.~Kunic, and J.~Tsotsos, ``Pie: A large-scale dataset and models for pedestrian intention estimation and trajectory prediction,'' in {\em 2019 IEEE/CVF International Conference on Computer Vision (ICCV)}, pp.~6261--6270, 2019.

\bibitem{nuscenes2019}
H.~Caesar, V.~Bankiti, A.~H. Lang, S.~Vora, V.~E. Liong, Q.~Xu, A.~Krishnan, Y.~Pan, G.~Baldan, and O.~Beijbom, ``nuscenes: A multimodal dataset for autonomous driving,'' {\em arXiv preprint arXiv:1903.11027}, 2019.

\bibitem{9709630}
S.~Ettinger, S.~Cheng, B.~Caine, C.~Liu, H.~Zhao, S.~Pradhan, Y.~Chai, B.~Sapp, C.~Qi, Y.~Zhou, Z.~Yang, A.~Chouard, P.~Sun, J.~Ngiam, V.~Vasudevan, A.~McCauley, J.~Shlens, and D.~Anguelov, ``Large scale interactive motion forecasting for autonomous driving : The waymo open motion dataset,'' in {\em 2021 IEEE/CVF International Conference on Computer Vision (ICCV)}, pp.~9690--9699, 2021.

\bibitem{Argoverse}
M.-F. Chang, J.~W. Lambert, P.~Sangkloy, J.~Singh, S.~Bak, A.~Hartnett, D.~Wang, P.~Carr, S.~Lucey, D.~Ramanan, and J.~Hays, ``Argoverse: 3d tracking and forecasting with rich maps,'' in {\em Conference on Computer Vision and Pattern Recognition (CVPR)}, 2019.

\bibitem{Houston2020OneTA}
J.~L. Houston, G.~C.~A. Zuidhof, L.~Bergamini, Y.~Ye, A.~Jain, S.~Omari, V.~I. Iglovikov, and P.~Ondruska, ``One thousand and one hours: Self-driving motion prediction dataset,'' in {\em Conference on Robot Learning}, 2020.

\bibitem{yu2020bdd100k}
F.~Yu, H.~Chen, X.~Wang, W.~Xian, Y.~Chen, F.~Liu, V.~Madhavan, and T.~Darrell, ``Bdd100k: A diverse driving dataset for heterogeneous multitask learning,'' in {\em Proceedings of the IEEE/CVF conference on computer vision and pattern recognition}, pp.~2636--2645, 2020.

\bibitem{9880320}
T.~Sun, M.~Segu, J.~Postels, Y.~Wang, L.~V. Gool, B.~Schiele, F.~Tombari, and F.~Yu, ``Shift: A synthetic driving dataset for continuous multi-task domain adaptation,'' in {\em 2022 IEEE/CVF Conference on Computer Vision and Pattern Recognition (CVPR)}, (Los Alamitos, CA, USA), pp.~21339--21350, IEEE Computer Society, jun 2022.

\bibitem{cabon2020virtualkitti2}
Y.~Cabon, N.~Murray, and M.~Humenberger, ``Virtual kitti 2,'' 2020.

\bibitem{long2016unsupervised}
M.~Long, H.~Zhu, J.~Wang, and M.~I. Jordan, ``Unsupervised domain adaptation with residual transfer networks,'' {\em Advances in neural information processing systems}, vol.~29, 2016.

\bibitem{tzeng2017adversarial}
E.~Tzeng, J.~Hoffman, K.~Saenko, and T.~Darrell, ``Adversarial discriminative domain adaptation,'' in {\em Proceedings of the IEEE conference on computer vision and pattern recognition}, pp.~7167--7176, 2017.

\bibitem{ghifary2015domain}
M.~Ghifary, W.~B. Kleijn, M.~Zhang, and D.~Balduzzi, ``Domain generalization for object recognition with multi-task autoencoders,'' in {\em Proceedings of the IEEE international conference on computer vision}, pp.~2551--2559, 2015.

\bibitem{ultralytics2023yolov8}
{Ultralytics}, ``Yolov8 documentation,'' 2023.

\bibitem{ultralytics2023rtdetr}
{Ultralytics}, ``Rt-detr: Real-time detection transformer,'' 2023.

\bibitem{ultralytics2023yolon}
{Ultralytics}, ``Yolo-nas documentation,'' 2023.

\end{thebibliography}
